\documentclass[letterpaper, 10 pt, conference]{ieeeconf}

\IEEEoverridecommandlockouts %
\usepackage{amsmath} %
\usepackage{amssymb} %
\usepackage{amsfonts}
\usepackage[sort,compress]{cite}
\usepackage[usenames,dvipsnames]{xcolor}
\usepackage{graphicx}
\usepackage{tabularx}
\usepackage{multirow}
\usepackage{mathtools}
\usepackage{esvect} %
\usepackage[binary-units=true]{siunitx}
\usepackage{caption}
\usepackage[pdftex, pdfstartview={FitV}, pdfpagelayout={TwoColumnLeft},bookmarksopen=true,plainpages = false, colorlinks=true, linkcolor=black, citecolor = black, urlcolor = black,filecolor=black , pagebackref=false,hypertexnames=false, plainpages=false, pdfpagelabels ]{hyperref}
\usepackage[T1]{fontenc} %
\usepackage{mathtools, cuted} %

\usepackage{balance} %
\usepackage{booktabs} %
\usepackage[export]{adjustbox} %

\newcolumntype{x}[1]{%
>{\raggedleft\hspace{0pt}}p{#1}}%

\usepackage{algorithm}
\usepackage[noend]{algpseudocode} %
\definecolor{commentclr}{RGB}{34, 139, 34}

\usepackage{tikz}

\usepackage{subcaption}
\DeclareCaptionLabelSeparator{periodspace}{.\quad}
\usepackage{makecell}

\usepackage{amsthm}

\theoremstyle{definition}

\usepackage{letltxmacro}
\LetLtxMacro\orgvdots\vdots
\LetLtxMacro\orgddots\ddots

\makeatletter
\DeclareRobustCommand\vdots{%
	\mathpalette\@vdots{}%
}
\newcommand*{\@vdots}[2]{%
	\sbox0{$#1\cdotp\cdotp\cdotp\m@th$}%
	\sbox2{$#1.\m@th$}%
	\vbox{%
		\dimen@=\wd0 %
		\advance\dimen@ -3\ht2 %
		\kern.5\dimen@
		\dimen@=\wd2 %
		\advance\dimen@ -\ht2 %
		\dimen2=\wd0 %
		\advance\dimen2 -\dimen@
		\vbox to \dimen2{%
			\offinterlineskip
			\copy2 \vfill\copy2 \vfill\copy2 %
		}%
	}%
}
\DeclareRobustCommand\ddots{%
	\mathinner{%
		\mathpalette\@ddots{}%
		\mkern\thinmuskip
	}%
}
\newcommand*{\@ddots}[2]{%
	\sbox0{$#1\cdotp\cdotp\cdotp\m@th$}%
	\sbox2{$#1.\m@th$}%
	\vbox{%
		\dimen@=\wd0 %
		\advance\dimen@ -3\ht2 %
		\kern.5\dimen@
		\dimen@=\wd2 %
		\advance\dimen@ -\ht2 %
		\dimen2=\wd0 %
		\advance\dimen2 -\dimen@
		\vbox to \dimen2{%
			\offinterlineskip
			\hbox{$#1\mathpunct{.}\m@th$}%
			\vfill
			\hbox{$#1\mathpunct{\kern\wd2}\mathpunct{.}\m@th$}%
			\vfill
			\hbox{$#1\mathpunct{\kern\wd2}\mathpunct{\kern\wd2}\mathpunct{.}\m@th$}%
		}%
	}%
}
\makeatother

\let\oldnl\nl%
\newcommand{\nonl}{\renewcommand{\nl}{\let\nl\oldnl}}%
\makeatother

\def\bn{\mathbb N}

\def\br{\mathbb R}

\def\sl{\mathcal{L}}

\newcommand\eqdef{\mathrel{\overset{\makebox[0pt]{\mbox{\normalfont\tiny def}}}{=}}}

\graphicspath{{figures/}}
\title{\LARGE \bf
View-Invariant Localization using Semantic Objects \\ in Changing Environments
	}

\author{Jacqueline Ankenbauer, Kaveh Fathian, Jonathan P. How%
	\thanks{J.\ Ankenbauer, K.\ Fathian and J.\ P.\ How are with the Department of Aeronautics and Astronautics, Massachusetts Institute of Technology.
	    {\{jpedlow, kavehf, jhow\}@mit.edu.} }
    \thanks{Supported by The Boeing Company under Cooperative Agreement MRA\#2017-656, ARL DCIST under Cooperative Agreement W911NF-17-2-0181, and by UPenn~ under ONR Award 584551.}
}%

\begin{document}

\maketitle
\thispagestyle{plain}
\pagestyle{plain}

\begin{abstract}
This paper proposes a novel framework for real-time localization and egomotion tracking of a vehicle in a reference map.
The core idea is to map the semantic objects observed by the vehicle and register them to their corresponding objects in the reference map. 
While several recent works have leveraged semantic information for cross-view localization, the main contribution of this work is a \textit{view-invariant} formulation that makes the approach directly applicable to any viewpoint configuration for which objects are detectable. 
Another distinctive feature is \textit{robustness to changes} in the environment/objects due to a data association scheme suited for extreme outlier regimes (e.g., 90\% association outliers). 
To demonstrate our framework, we consider an example of localizing a ground vehicle in a reference object map using only cars as objects.
While only a stereo camera is used for the ground vehicle, we consider reference maps constructed a priori from ground viewpoints using stereo cameras and Lidar scans, and georeferenced aerial images captured at a different date to demonstrate the framework's robustness to different modalities, viewpoints, and environment changes.
Evaluations on the KITTI dataset show that over a $3.7$\,km trajectory, localization occurs in $36$\,sec and is followed by real-time egomotion tracking with an average position error of $8.5$\,m in a Lidar reference map, and on an aerial object map where $77\%$ of objects are outliers, localization is achieved in $71$\,sec with an average position error of $7.9$\,m.
\end{abstract}

\section{INTRODUCTION}\label{sec:intro}

Localization is the process of determining the position and orientation (i.e., pose) of a vehicle in its environment.
As global navigation satellite systems can be denied (e.g., in subterranean environments), 
localization based on onboard sensor observations of the surrounding environment is of great interest.   
Given a reference map of the environment, the key challenge of localization is to correctly associate vehicle observations to their correspondence in the reference map. 
This difficulty can be caused by 1) different perspectives/viewpoints between the reference and vehicle maps (e.g., localizing a ground vehicle in an aerial map is challenging since the same location is visually different from the ground and aerial views), 2) changes in the environment (e.g., lighting differences, seasonal changes, and objects being created, moved, or removed causes a discrepancy between reference and vehicle maps), 3) different sensing modalities (e.g., localizing a camera image in a Lidar point cloud is challenging due to different data attributes).

\begin{figure}[t!]
	\centering
	\includegraphics[width=1.0\columnwidth]{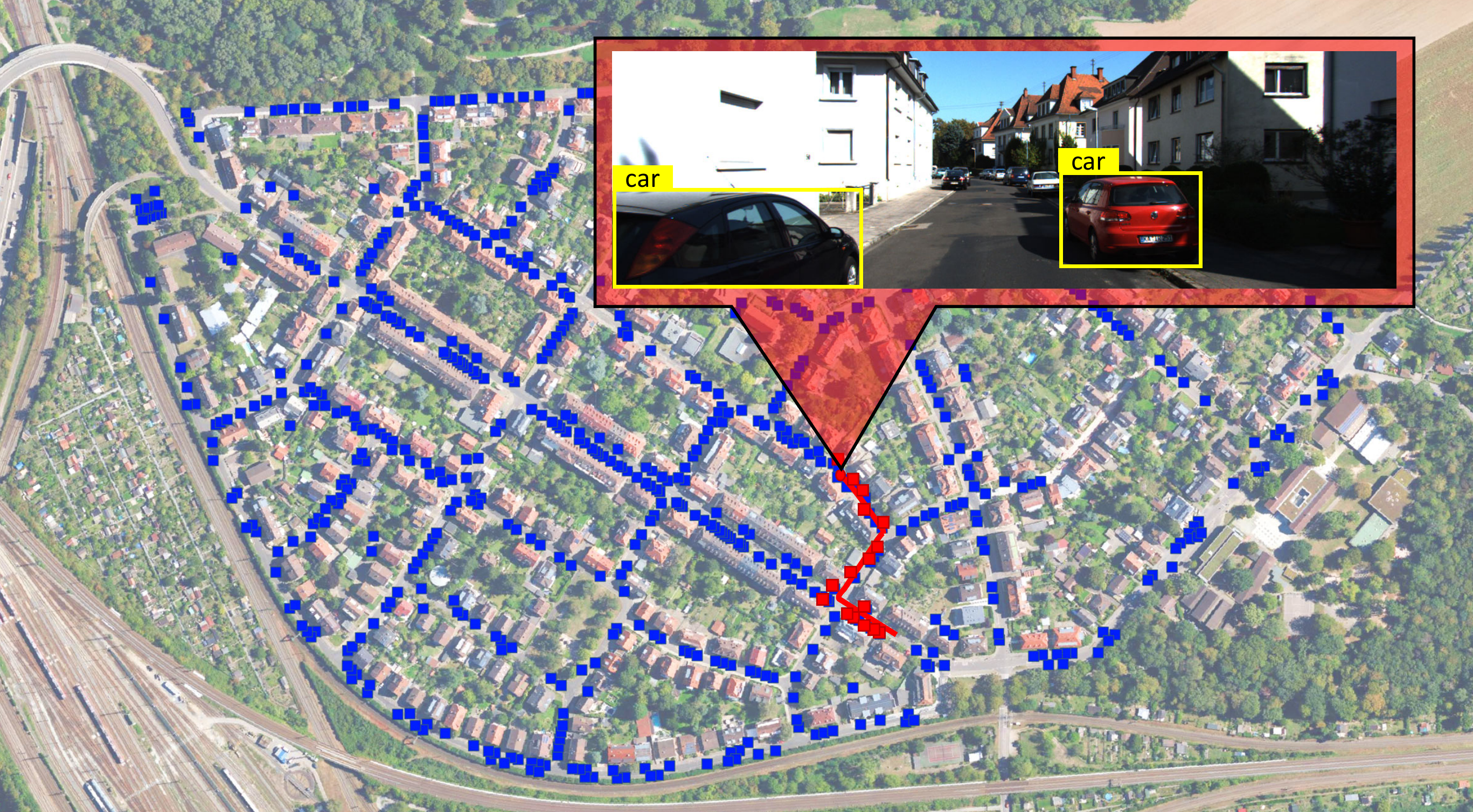}
	\caption{
	Localization and tracking the pose of a ground vehicle 
	in a reference object map (blue squares) built from cars in a georeferenced aerial image. 
	The ground vehicle is localized by creating an object map of the cars observed while driving (red squares) and registering them with the cars in the reference map.
	Over a $3.7$\,km trajectory, our framework localizes the vehicle after $476$\,m (in $71$\,sec), and tracks it henceforth with an average position error of $7.9$\,m. This is despite the ground vehicle dataset and the aerial image being captured in different years and having nonidentical cars with only $23\%$ (roughly) at the same location, and the reference map having an accuracy of (at best) $2$\,m.
	}
	\label{fig:air_ground_example}
	\vspace*{-0.3em}
\end{figure}

Toward addressing the above three challenges, this work presents a novel framework based on registering semantic objects observed by the vehicle (vehicle map) to the objects in the environment (reference map).
The proposed framework is view-invariant because it uses a global registration technique that can match objects in any configuration as long as objects are observable/detectable in the vehicle and reference views.
Additionally, the chosen data association scheme makes the framework highly robust to association outliers. Therefore, objects that are dynamic, missed, misclassified, or represented poorly are treated as outliers and rejected--even if they constitute a great portion of the data--without affecting the localization. Further, using objects for localization introduces a degree of robustness to environment appearance changes because object classifiers can be trained to accurately identify objects in different lighting and seasons.

\begin{figure}[b]
	\centering
	\includegraphics[width=0.99\columnwidth]{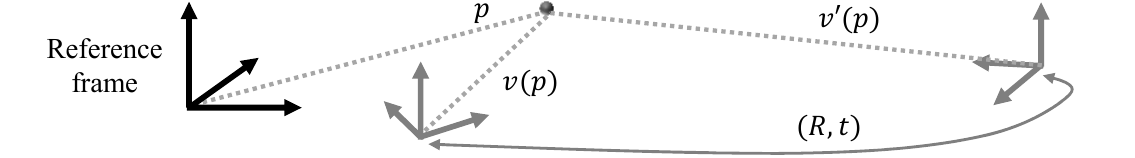}
	\caption{
		A localization framework $f$ is \textit{view-invariant} if and only if for all transforms $(R,t)$ between the viewpoints $v(p)$ and $v'(p)$ of the point $p$ in the reference frame, it holds that $f(v(p)) = f(v'(p)) = p$.
	}
	\label{fig:view_invar}
\end{figure}

Objects provide a common framework for localization using different sensing modalities. For example, modern classifiers can detect the same object classes in Lidar point clouds (e.g, PointNet \cite{qi2017pointnet}, VoxelNet \cite{zhou2018voxelnet}, RangeNet++ \cite{milioto2019rangenet++}) and in images (e.g., YOLO \cite{redmon2016you}, Mask R-CNN \cite{He_2017_ICCV}). By registering the objects observed in these modalities, a camera can be localized in a Lidar map.
In this work, each observed object is represented by a 3D point and its class. This compact representation leads to a small map size for a large environment, and consequently enables more efficient communication of the reference and vehicle maps (this is crucial in a multi-agent setting with bandwidth constraints).

We demonstrate the framework experimentally for the case of localizing a ground vehicle in a reference map, using only cars as semantic objects.
Only stereo camera images from the KITTI dataset \cite{geiger2012cvpr} are used to create the ground vehicle's object map.  We consider reference maps constructed from cars seen a priori from ground viewpoints using stereo cameras and Lidar scans, and aerial maps captured on different dates to demonstrate the framework's robustness to viewpoints, sensing modalities, and environment changes (see Fig.~\ref{fig:air_ground_example}).
The localization performance is analyzed in terms of pose accuracy and convergence time.  With accuracy comparable to state-of-the-art methods, our framework is also view-invariant and robust to environment changes.

In summary, the contributions of this work are:
\begin{itemize}
    \item \textit{\textbf{View-invariant}} formulation of localization as the registration of semantic objects, applicable to all cross-view scenarios with jointly observed objects (see Fig.~\ref{fig:view_invar}). 
    \item Resilient to \textit{\textbf{environment/object changes}} due to a robust data association framework capable of handling large outlier ratios in associations (>90\%).     
    \item Demonstrations on real data from the KITTI dataset showing comparable accuracy to state-of-the-art methods, cross-modal localization between Lidar and stereo maps using objects, and geo-localization in an aerial map captured on a different date with significantly changed objects. 
\end{itemize}
While prior work may share a subset of above contributions, what distinguishes this systems paper is their combination (as discussed in related work).

\section{Related Work}

The large body of work on localization has close ties to loop closure detection, place recognition, and image retrieval literature. We focus our review on the most recent/related work in these domains.

\subsection{Appearance-Based Methods}

Appearance-based methods use images for localization via finding the most visually similar image in the reference database to the locally captured image. 
The visual similarity is assessed based on low-level information such as color
and reflectance values \cite{veronese2015re,vora2020aerial} or (low-level) visual features/descriptors
\cite{senlet2011framework,noda2010vehicle}.
Early works such as \cite{chen2011city}, \cite{li2010location} use local feature descriptors to compare and match images taken from different perspectives. 
Majdik et al. \cite{majdik2013mav} use such features with simulated images from Google Street View to match against images from a quadrotor flying through an urban environment.
Methods based on low-level features are impacted the most by changes in the environment such as illumination or seasonal changes, and, more importantly, fail under extreme viewpoint difference between the vehicle and reference images (e.g., aerial-ground).

\subsection{Cross-View Localization}
The term ``cross-view'' is mainly used by the literature for geo-localization of a ground vehicle in an aerial/satellite map. However, some works also consider localization of an aerial vehicle (e.g., a drone) in the satellite image \cite{wang2021each}.
Current learning-based state-of-the-art \cite{shi2019spatial,rodrigues2021these,wang2021each} use Siamese network architecture \cite{kim2017satellite,tian2020cross},
and can return a coarse localization (accuracy of hundreds of meters) across a very large area (e.g., city-wide).
When coupled with particle filters, these techniques can provide a higher accuracy (tens of meters) in geo-tracking applications \cite{downes2022city}.
Due to the air-ground view, cross-view algorithms are generally robust to viewpoint variations, however, they are not necessarily view-invariant (see definition in Fig.~\ref{fig:view_invar}). 
For example, even though these methods work for the extreme air-ground view, they may not work directly for other views (e.g., ground-ground).
Further, many are relatively slow for real-time robotic applications.

Air-ground localization can be achieved by other techniques, such as \cite{wolcott2014visual,gawel20173D,barsan2020learning} that can obtain highly accurate localization (centimeter-level) when a high-definition point cloud map of the environment is available. 
While accurate, these methods do not work with image maps and require  dense point clouds.

\subsection{Semantic-Aided Methods}

Several recent works have leveraged semantic information for localization. Different types of semantic information used in the air-ground localization literature include building outlines \cite{vysotska2017improving,kim2019fusing}, general vertical structures \cite{kummerle2011large,wang2017flag, wang2022ltsr}, lane markings \cite{pink2008visual,javanmardi2017towards}, binary ground-nonground distinction \cite{viswanathan2016vision}, vehicle trajectory \cite{brubaker2013lost}, \cite{floros2013openstreetslam}, semantic segmentation \cite{noda2010vehicle,miller2021any},  
buildings \cite{matei2013image,senlet2014hierarchical,tian2017cross},
intersections and building gaps \cite{yan2019global},
and 3D labeled points \cite{stenborg2018long}, \cite{liu2019global}.
While semantic classifiers are typically robust to viewpoint variations and to seasonal and lighting changes, using semantics limits the applicability to environments where the desired classes are present (e.g., urban).

\begin{figure*}[h!]
	\centering
	\begin{subfigure}[b]{0.990\textwidth}\includegraphics[trim = 0mm 0mm 0mm 0mm, clip, width=1\textwidth] {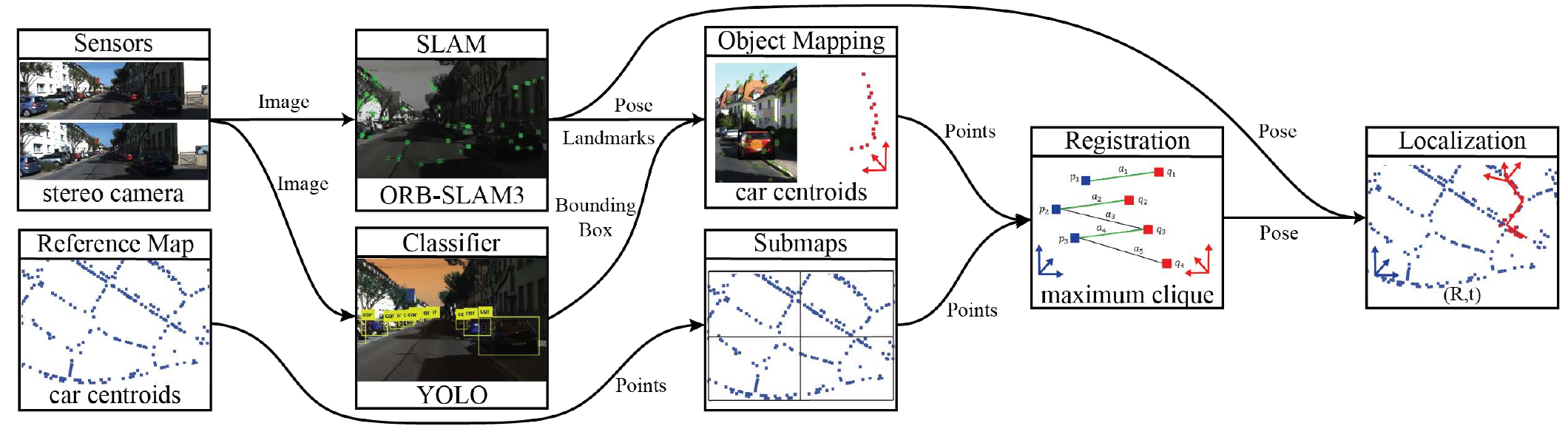}
	\end{subfigure}%
	\caption{Main components of the tracking and localization framework.  From onboard sensors, the simultaneous localization and mapping (SLAM) system estimates the vehicle's pose in its local frame, the classifier identifies objects, and the vehicle object map is built.  The objects in the reference map are grouped into submaps and the vehicle is localized by performing data association to align the two maps. }
	\vspace*{-1.5em}
	\label{fig:block_diagram}	
\end{figure*}

\subsection{Placement of This Work}
This work leverages semantic information, however, its view-invariant formulation and ability to handle changes in the environment (in terms of objects) distinguishes it from prior art.
Methods that leverage semantic information for localization have not shown to be robust to changes in the semantic data nor completely view-invariant.  Methods are typically designed specifically for ground-ground, air-air, or air-ground localization, and the ability of one framework to handle all cases is rarely analyzed.
A subtler difference is that instead of one-time localization, the framework is designed for real-time continuous localization and tracking of the vehicle's pose in the environment, which is important for robotic applications.
Only few works \cite{miller2021any, fervers2022continuous, downes2022city} have considered this case.

\section{PIPELINE} \label{sec:framwork}

Components of the proposed pipeline are shown in Fig.~\ref{fig:block_diagram} and explained in detail below. 

\subsection{Onboard Sensors}
Sensor measurements are used to estimate the vehicle's motion and to detect and reconstruct objects.
Any sensor suite that can provide egomotion estimation, object detection, and depth can be used (e.g., camera+IMU, Lidar); in this work we experiment with a stereo camera.

\subsection{Reference Map and Submaps}

The pipeline takes in a reference map of objects in the environment, which can be fully constructed a priori or changing in real-time (e.g., dynamic reference maps are important in multi-agent settings).  Our experiments were conducted on static maps.
Objects can be of any type/class provided that a small subset of objects detected by the vehicle are co-located with the objects in the reference map, so they can be registered and thus provide localization.

Each object is represented by its 3D centroid $p \in \br^3$ and an integer type/class $c\in \bn$.
Therefore, the reference map can be denoted as a set of tuples 
$S_{\text{ref}} \eqdef \{(p_i,c_i):\, p_i \in \br^3,\,c_i\in \bn, i=1,\dots, n\}$.
Since the computational complexity of the registration module grows with the number of objects (Sec.~\ref{subsec:registration}), to achieve real-time registration, the reference map can be divided into $k$ smaller submaps $S^i_{\text{ref}} \subset S_{\text{ref}}$, $i = 1, \dots, k$.
The submaps may overlap and they must cover the entire reference map (i.e., $\bigcup_i S^i_{\text{ref}}  = S_{\text{ref}}$).

\subsection{SLAM System}\label{subsec:slam-system}
The purpose of the SLAM module is to provide odometry estimates for the vehicle and to reconstruct a point cloud map of the environment.
Our implementation uses ORB-SLAM3 \cite{campos2021orbslam}, which takes in the stereo camera images as input and returns the vehicle's pose and 3D reconstructed ORB feature points. 
This output is used subsequently to reconstruct the objects and track the egomotion of the vehicle in the reference map.

\subsection{Classifier} \label {subsec:obj-detection}
The role of the classifier is to detect semantic objects of interest in the sensor data. 
We only consider cars as objects in our demonstrations, however, other types of objects can be used instead (e.g., traffic signs, trees).
Provided an RGB image as input, we use YOLO-V3 \cite{bjelonicYolo2018} to return a bounding box and class for each detected object. This bounding box is then used to estimate the object's centroid.

\begin{figure*}[th!]
	\centering
	\begin{subfigure}[b]{0.333\textwidth}\includegraphics[trim = 0mm 0mm 0mm 0mm, clip, width=0.99\textwidth] {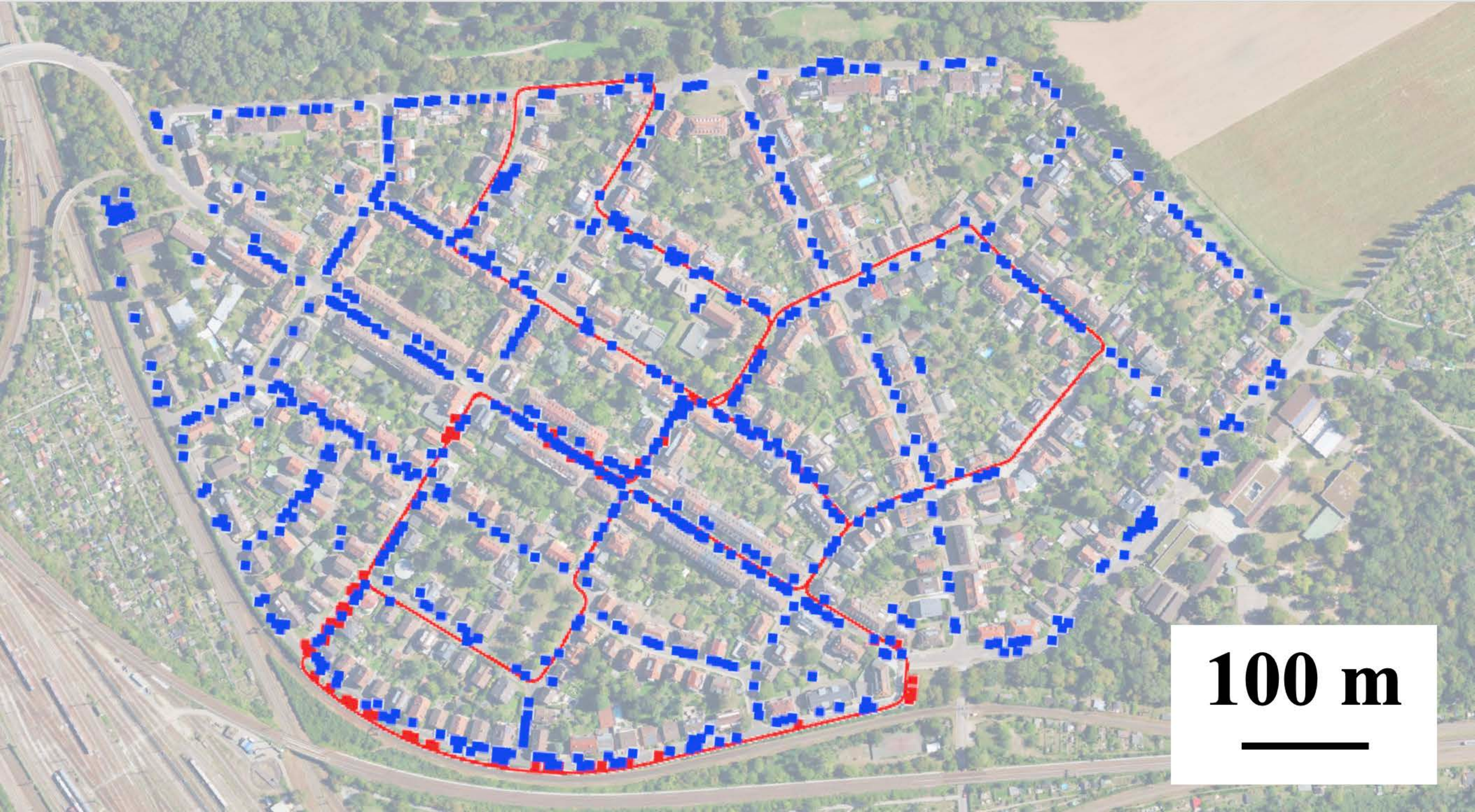}
	\end{subfigure}%
	\begin{subfigure}[b]{0.333\textwidth}\includegraphics[trim = 0mm 0mm 0mm 0mm, clip, width=0.99\textwidth] {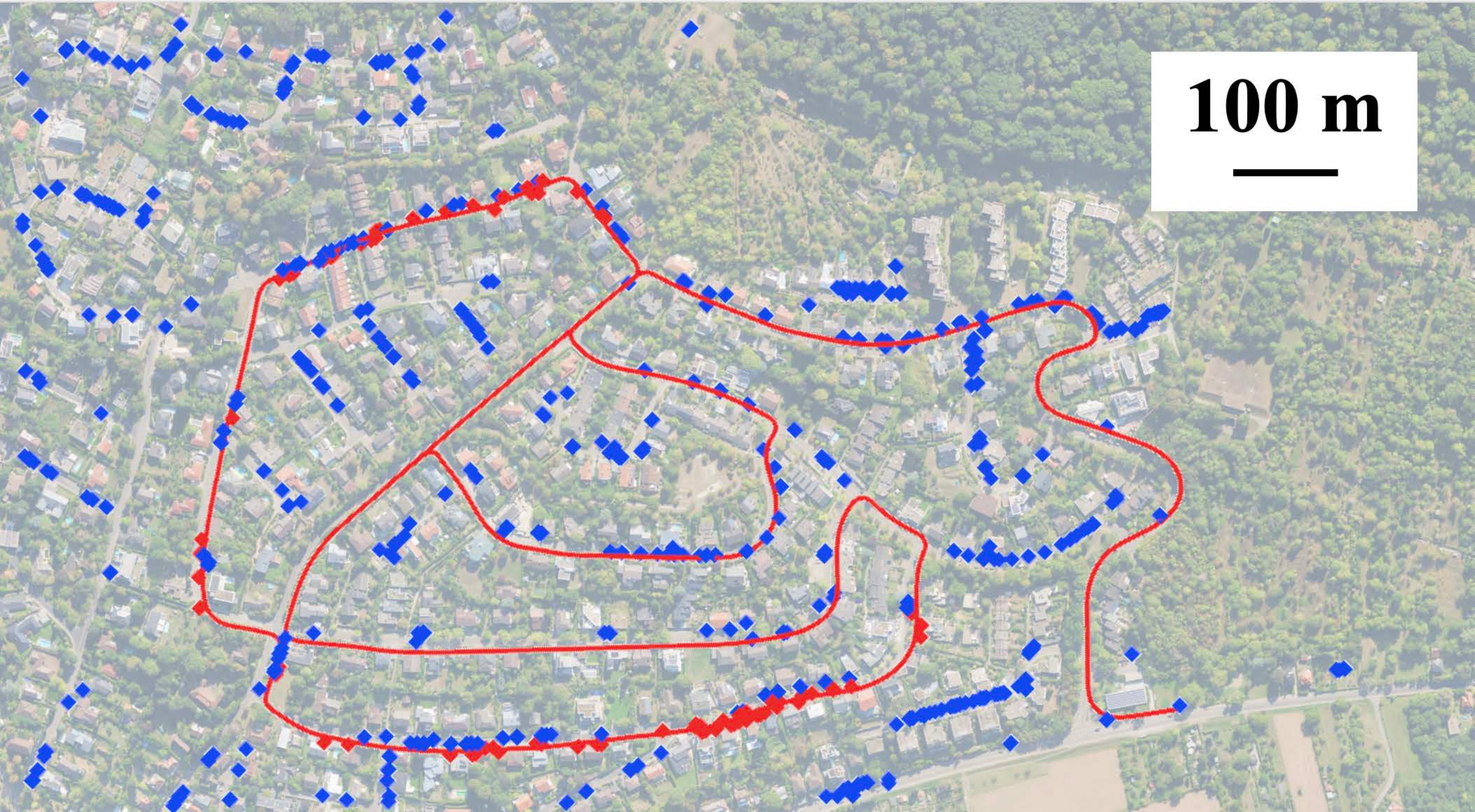}
	\end{subfigure}%
	\begin{subfigure}[b]{0.333\textwidth}\includegraphics[trim = 0mm 0mm 0mm 0mm, clip, width=0.99\textwidth] {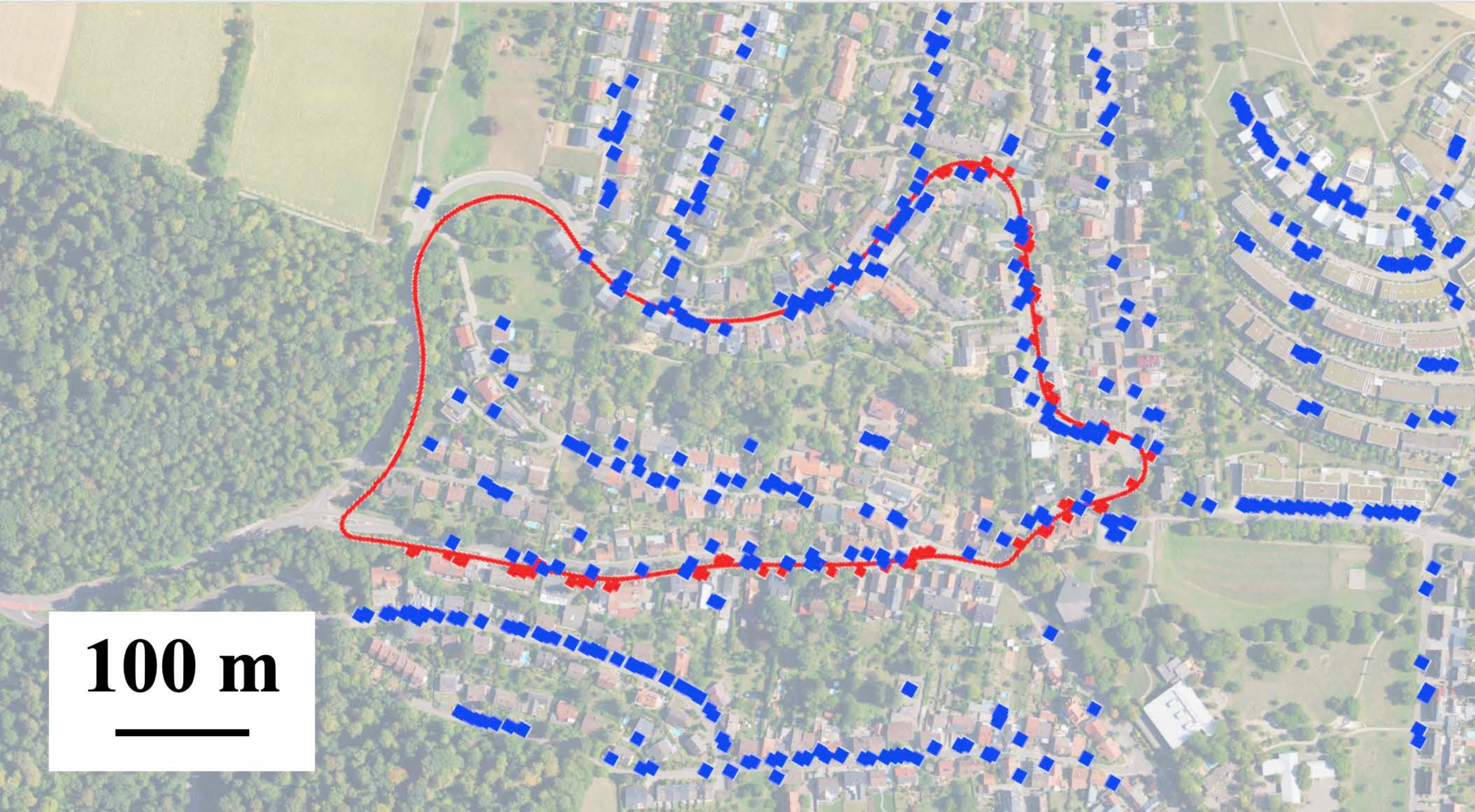}
	\end{subfigure}%
	\caption{
	(Left to right) KITTI Sequences 0, 2, and 9. 
	The localized trajectory of the ground vehicle (red line), vehicle map (red squares), and reference map (blue squares) constructed from the background orthophotos captured on a different date are shown.
	 The background images are provided for reference and only the cars (blue and red squares) are used for registration. }
	\vspace*{-1.5em}
	\label{fig:aerial_localization}	
\end{figure*}

\subsection{Object Mapping} \label{subsec:mapper}

An object's centroid is estimated using the 3D landmarks reconstructed by the SLAM system that correspond to the object, meaning the 3D reconstructed landmarks that project inside the object's bounding box in the image.
We approximate the distance to the object from the current vehicle pose by averaging the distance to all corresponding ORB landmarks.
Given this distance, we represent the object centroid with the 3D point that is at the computed distance and projects onto the \textit{center} of the bounding box.
Subsequently, the vehicle object map is constructed by tracking all object centroids across consecutive camera frames and fusing them into one map.  
We denote the vehicle object map by the set of 3D points and their classes as ${S_{\text{loc}} \eqdef \{(q_i,c_i):\, q_i \in \br^3,\, c_i\in \bn, i=1,\dots, m \}}$.
Instead of using the entire vehicle map, a subset of the recently observed objects is used for localization. This speeds up the registration by reducing the problem size and reduces the registration errors that are due to the accumulated drift in the SLAM trajectory.

\subsection{Registration} \label{subsec:registration}

Robust registration is the core component of the proposed framework. 
We use a graph-based formulation to solve the registration problem by finding the largest set of geometrically consistent objects that match between the reference submaps and vehicle map.
Denoting by $a_i =(p_i, q_i)$ the association that matches the points $p_i$ and $q_i$, two associations $a_i$ and $a_j$ are considered \textit{geometrically consistent} if and only if the distance between the points is preserved, i.e., $\| p_i - p_j \| = \| q_i - q_j \|$.
In practice, due to noise and inaccuracies, a threshold $\epsilon$ is set in order to consider associations consistent when $d(a_i, a_j) \eqdef | \, \| p_i - p_j \| - \| q_i - q_j \| \,| < \epsilon$. 
Now, by denoting the set of associations between objects of the same class in the reference and vehicle maps as
${A \eqdef \{ (o_i, o_j ): \, o_i \in S_{\text{ref}},\, o_j \in S_{\text{loc}},\, \text{class}(o_i) = \text{class}(o_j) \}}$, the problem of finding the largest set of consistent associations, $A^*_{\text{c}}$, can be defined formally as 
\begin{gather} \label{eq:maxclique}
	\begin{array}{ll}
		\underset{A_{\text{c}} \subset A}{\text{maximize}} & | A_{\text{c}} |
		\\
		\text{subject to} & d(a_i, a_j) < \epsilon, ~ \forall_{a_i,a_j \in A_{\text{c}}}.
	\end{array}
\end{gather}

\begin{figure}[b]
	\centering
	\includegraphics[width=0.99\columnwidth]{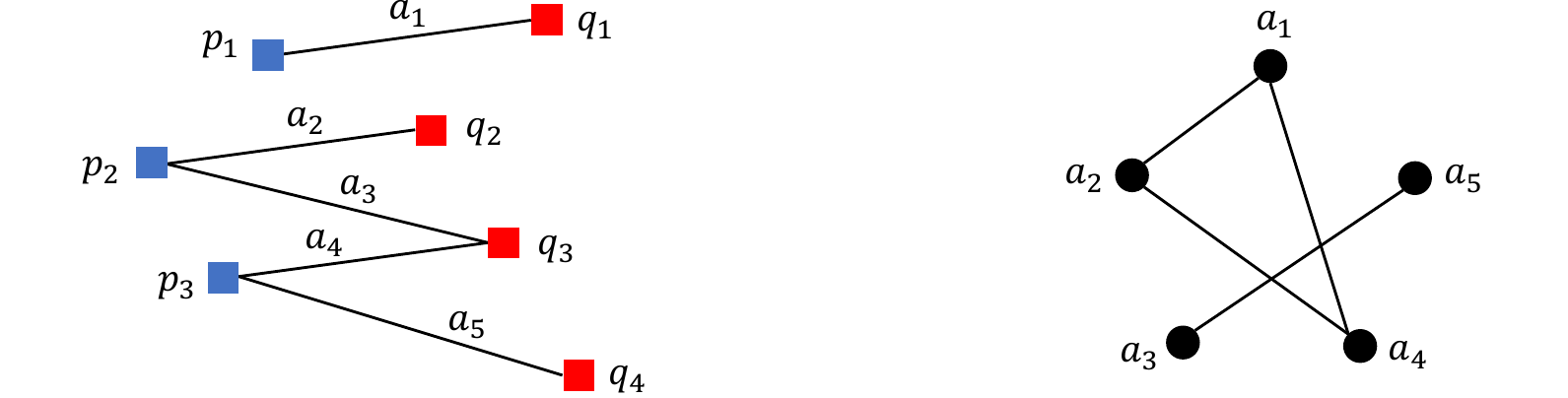}
	\caption{Maximum clique formulation for registration. (Left) Points $p$ and $q$ are matched by associations $a$. (Right) 
	Graph with nodes representing associations and edges indicating their consistency (i.e., associations with identical distances between their endpoints).
	The largest clique ${A^*_{\text{c}} = \{a_1,a_2,a_4\}}$ is the largest set of consistent associations.}
	\label{fig:max_clique}
\end{figure}

Problem \eqref{eq:maxclique} can be modeled as a graph with associations representing the graph nodes and edges representing consistent associations. The optimal solution of \eqref{eq:maxclique} is equivalent to the maximum clique of the graph. 
Fig.~\ref{fig:max_clique} illustrates this point. 
Finding the maximum clique is, in general, an NP-hard problem, which implies that exact algorithms become computationally inhibiting as the graph size grows.
However, for sparse graphs (resulting from many inconsistent associations created by an all-to-all scheme), the problem can be solved relatively quickly using the parallel maximum clique (PMC) algorithm \cite{rossi2013parallel}.

We use an all-to-all association scheme, i.e., we initially associate an object in the reference map to every object in the vehicle map (of the same class) and vice versa.
Despite the large number of association outliers generated by the all-to-all association and by inconsistent points, solving 
\eqref{eq:maxclique} using PMC is guaranteed to return the maximum set of consistent associations.
The all-to-all scheme ensures that geometrically consistent objects are matched from any configuration, hence the \textit{\textbf{view-invariant}} claim. 
The least-square fitting of matched objects using Arun's method \cite{arun1987least} gives the optimal transformation $(R,t) \in \text{SO}(3)$ that registers the objects.

\subsection{Localization} \label{subsec:localize}

The final module in the pipeline is localization.
This module continuously tracks the vehicle's pose in the reference map by transforming the vehicle's pose in the SLAM frame to the reference frame using the latest registration results.
The highest number of matched points in all previous registrations is kept and a new registration is only accepted when the number of matched points is 90\% or higher than the highest quality registration from the history.

\section{EXPERIMENTAL EVALUATIONS} \label{sec:expr}

We demonstrate the framework experimentally by localizing a ground vehicle in a reference map. 
We use stereo camera images from the KITTI dataset \cite{geiger2012cvpr} to create the ground vehicle's object map, as this dataset is used by several prior works and provides a benchmark for comparison.
The framework is implemented in C++ and modules use the robot operating system (ROS)~\cite{quigley2009ros} to communicate. 
The pipeline runs real-time on a laptop with an i7-8750H CPU, 16 GB RAM, and Nvidia 1060 GPU.
The objectives are to evaluate our framework with different map modalities, different viewpoints, environment/object changes, and to perform an ablation study of errors.

\subsection{Evaluation Scenarios}

To create a particularly challenging environment, we only consider cars as semantic objects (as opposed to buildings \cite{miller2021any} or lane markings \cite{pink2008visual,javanmardi2017towards} considered in other work).
This enables us to test the framework with a significantly changed semantic object map because cars are dynamic or quasi-static (i.e., parked cars are at different locations on different dates).  It is possible to localize a vehicle in a reference map which contains none of the same cars as in the vehicle map because cars are a proxy for parking spaces and our framework is indifferent to visual car attributes (e.g., color, model) for registration.
We consider KITTI sequences numbered 0, 2, and 9 since they have been used by prior art as benchmarks \cite{miller2021any}. 
Algorithm parameters are tuned on Sequence 0, and the same values are used for other sequences.
In particular, we use a threshold of $\epsilon = 5$\,m  for registration (defined in Sec. \ref{subsec:registration}) and restrict the size of the vehicle object map to the last $75$ seen objects/cars to mitigate the effect of drift.
We consider registration results that match at least $20$ objects across the reference and vehicle maps, but this number may increase as more accurate registrations are found.  Due to the larger size of the reference map for KITTI Sequence 0, it is grouped into $4$ submaps with $50\%$ overlap, but no submaps are used for KITTI Sequences 2 and 9.

\begin{figure}[t!]
	\centering
	\includegraphics[width=0.99\columnwidth]{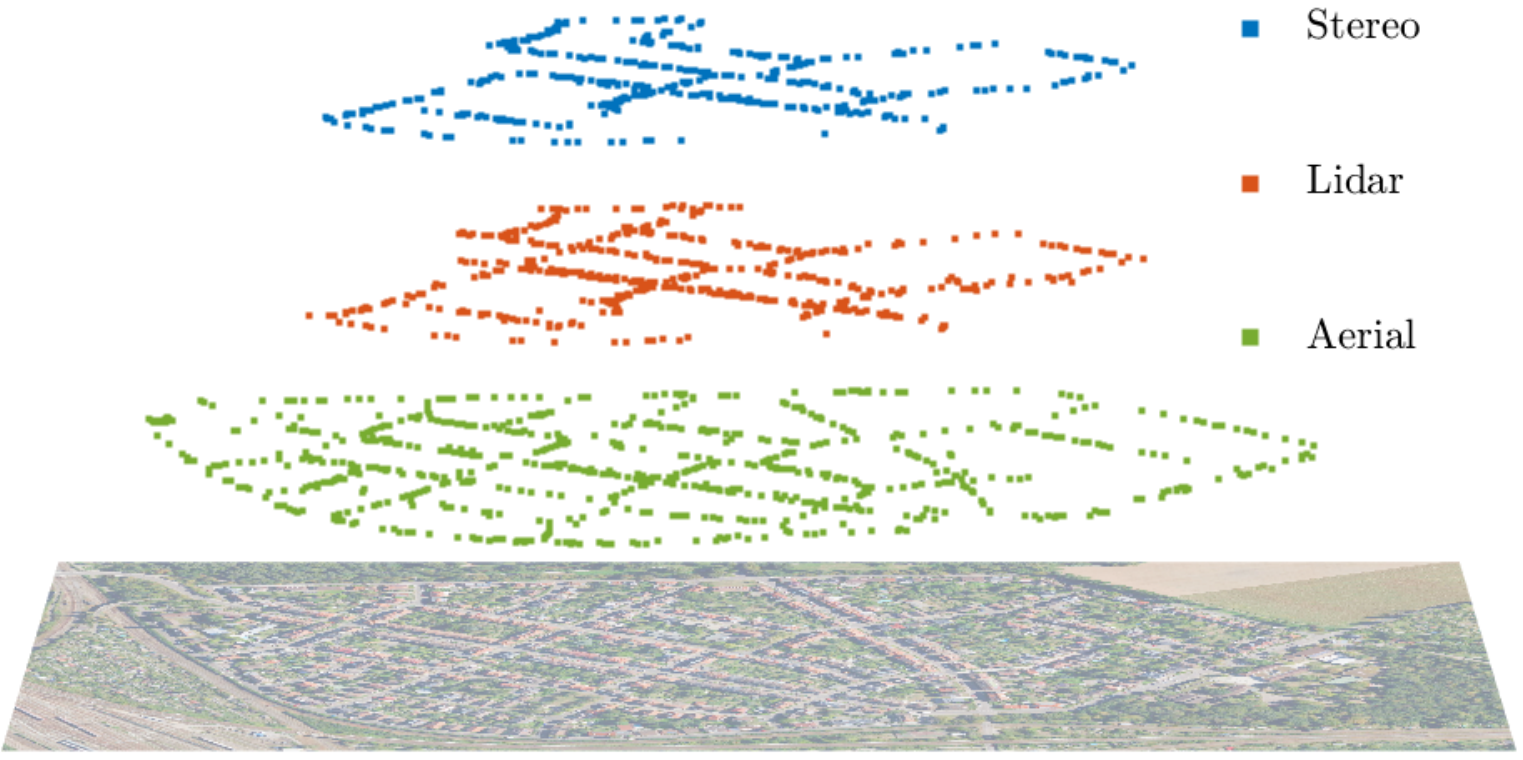}
	\vspace*{-0.3em}
	\caption{Reference object maps corresponding to KITTI Sequence 0 constructed using stereo camera images (ground view), Lidar scans (ground view), and Bing's georeferenced images (aerial view). Each square represents a car. The bottom image is for reference.}
	\label{fig:maps}
	\vspace*{-0.5em}
\end{figure}

For the reference map, we consider three scenarios of object/car maps created from 1) ground-view stereo camera images, 2) ground-view Lidar scans, and 3) aerial-view georeferenced images (orthophotos) to test the framework's robustness to viewpoints, sensing modalities, and environment changes that exist across these maps.
Fig.~\ref{fig:maps} shows an overlay of these object maps on the area corresponding to Sequence~0, where cars are represented by their centroids as 3D points.  The aerial reference map presents the most challenging case because it has the largest number of objects, the highest object outlier ratio, is constructed from a different viewpoint than the vehicle, and the orhtophotos were captured on a different date.
Fig.~\ref{fig:aerial_localization} illustrates snippets of the localization results in the aerial-ground case.

\subsection{Baseline} \label{subsec:outlier_effect}

To create a baseline case which factors out the effects of viewpoint, sensing modality, and environment changes, we localize the ground vehicle in a reference map built from the identical data (KITTI stereo images) as used by the ground vehicle. 
This reference map (shown in Fig.~\ref{fig:maps}) is created as described in Sec.~\ref{subsec:mapper}, but using the ground truth vehicle pose instead of the ORB-SLAM estimated pose.
While the reference map consists of $374$ cars, the vehicle's map has $437$ cars because the object mapping module misidentifies previously-seen objects as new instances due to drift in SLAM pose estimates.
The error in the pose estimates also affects the object centroid estimation.
For these reasons, despite identical sensing modalities and viewpoints, only $179$ of the $437$ cars seen by the vehicle match to a car in the reference map (based on the $\epsilon = 5$ consistency threshold). Thus, the vehicle map contains $59\%$ object outliers.

\begin{figure}[t!]
	\centering%
	\begin{subfigure}[b]{0.99\columnwidth}
		\centering
		\includegraphics[trim = 0mm 6mm 0mm 0mm, clip, width=1\textwidth]{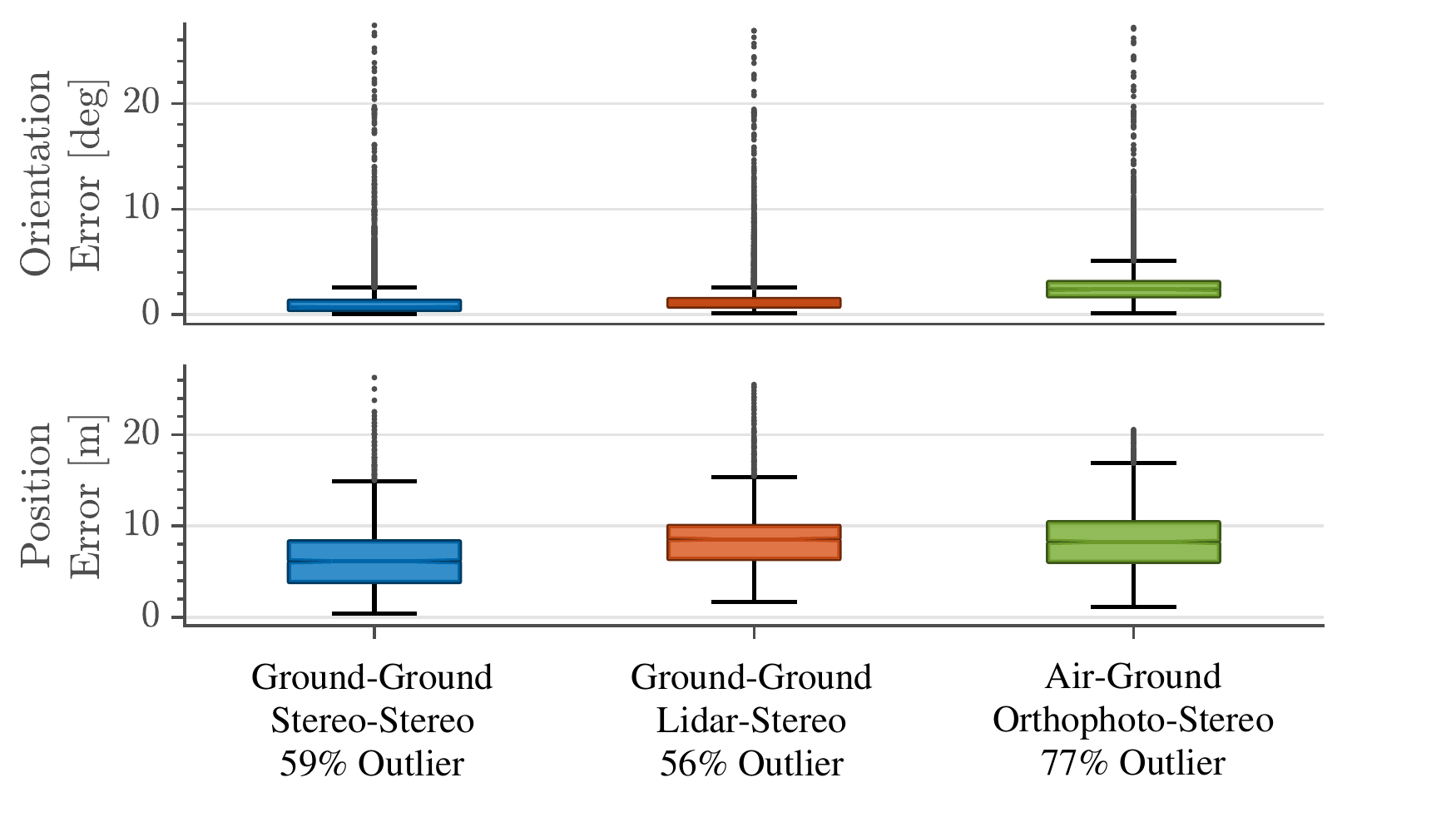}
	\end{subfigure}
	\vspace*{-0.3em}
	\caption{Error statistics for localization of the ground vehicle in KITTI Sequence 0 on three different reference maps (shown in Fig.~\ref{fig:maps}). Errors are comparable across cases, showing that our framework \textbf{is not affected by different object outlier percentages, map modalities, or viewpoints}.  The orientation errors have large tails due to temporary error spikes in ORB-SLAM orientation estimates during an acute turn in the trajectory.}
	\vspace*{-0.3em}
	\label{fig:boxplot}
\end{figure}

The localization time is defined as the time it takes from the start of the sequence until the vehicle finds a correct registration, which is $32$\,sec for this case.
The position and orientation errors are computed at each time step after the localization time via 
${e_{p}(t) \eqdef \parallel T_{\text{gt}}(t) - T(t) \parallel}$ and
${e_{o}(t) \eqdef \frac{180}{\pi} \arccos{(\mid <Q_{\text{gt}}(t), Q(t)> \mid)}}$,
where $T$ and $Q$ represent position and quaternion orientation, and ``gt'' denotes the KITTI ground truth.
Fig.~\ref{fig:boxplot} shows the error statistics.  
The pipeline successfully rejects outlier objects and obtains an average localization error of $6.1$\,m.

\subsection{Robustness to Map Modality}

To analyze the effect of map modality, an object map is created from the KITTI ground truth poses and the Lidar scans of a Velodyne Lidar mounted on the ground vehicle (shown in Fig.~\ref{fig:maps}).
Here, each car centroid in the reference map is computed as the median of the Lidar points associated with that car, as labeled in the SemanticKITTI dataset \cite{behley2019iccv}.
The vehicle map has an object outlier ratio of $56\%$ with respect to the Lidar map, which  consists of $445$ cars.
Since both the stereo baseline and Lidar reference maps have similar object outlier ratios ($59\%$ and $56\%$) and are constructed from the same ground truth poses, comparison of the localization errors isolates the effect of sensing modality. The error statistics are shown in Fig.~\ref{fig:boxplot}. The average position errors are $6.1$\,m and $8.5$\,m and the localization times are $32$\,sec and $36$\,sec for the stereo-stereo and stereo-Lidar cases, respectively.
The errors are comparable, which indicates robustness of the pipeline to sensing modalities.
The small difference in errors is due to the different centroid estimates for the same car in the Lidar and baseline reference maps as centroids are computed using different methods.

\begin{table*}[th!] %
\scriptsize
\centering
\caption{
	Statistics of KITTI ground vehicle localization in Bing's georeferenced aerial images using only cars as semantic objects.  Since the aerial reference map is 2D, the position and orientation errors are calculated after the pose estimate is projected onto the reference map plane.
}
\vspace*{-0.3em}
\setlength{\tabcolsep}{2.5pt}
\begin{tabular}{c c c c c c c c c c c c c c c c c c c c }
	\toprule
	\makecell{Reference-Vehicle\\ Viewpoint} & \makecell{KITTI \\ Sequence \\ $[$\#$]$} && \makecell{ Average \\ Position Error \\ $[$m$]$ } && \makecell{Average \\ Orientation Error \\ $[$deg$]$} && \makecell{Time to\\Localize \\$[$sec$]$ } && \makecell{Distance to\\  Localize\\ $[$m$]$} && \makecell{Trajectory\\ length\\ $[$m$]$} && \makecell{Objects to\\ Localize \\ $[$\#$]$}  && \makecell{Objects in \\ Reference Map \\ $[$\#$]$} && \makecell{Objects Seen \\ by Vehicle\\ $[$\#$]$} && \makecell{Object \\Outliers \\ $[$\%$]$} \\ 
	\toprule
	Aerial-Ground &  \makecell{$0$\\$2$\\$9$}   && \makecell{$7.9$\\$10.6$\\$10.9$} && \makecell{$1.93$\\$1.03$\\$1.08$} &&  \makecell{$71$\\$183$\\$108$}  &&  \makecell{$476$\\$1836$\\$1077$}  &&  \makecell{$3724$\\$5067$\\$1705$}  &&  \makecell{$45$\\$81$\\$56$}  &&  \makecell{$675$\\$409$\\$391$}  &&  \makecell{$418$\\$200$\\$97$}  &&  \makecell{$77$\\$85$\\$77$}  \\
			 
	\bottomrule
\end{tabular}
\vspace*{-1.3em}
\label{tbl:benchmarks}
\end{table*}

\subsection{Robustness to Viewpoint and Environment Changes}

To demonstrate that the framework is view-invariant, we consider air-ground localization using a reference map created from Bing aerial images \cite{bing_aerial_images}.
This map has a fixed scale since orthophotos are georeferenced with the EPSG:3857 coordinates, which provide an accuracy of (at best) 2 meters.
We annotated the cars manually in aerial images (which can be automated by classifiers trained for aerial/satellite images \cite{Ding_2019_CVPR, li2022oriented}).
The vehicle map has $77\%$ object outliers with respect to the aerial map, consisting of $675$ cars (shown in Fig.~\ref{fig:maps}). 
The localization error statistics are reported in Fig.~\ref{fig:boxplot}.

Compared to the stereo and Lidar maps, the aerial reference map provides the most challenging localization scenario because the orthophotos and the KITTI dataset are captured in different years and the cars are unlikely to be identical and/or at the same location.
Further, some cars could not be labeled in the aerial images due to
natural occlusions (e.g., trees, shadows) and low image resolution.
Despite the limited $2$\,m accuracy of the EPSG coordinates and challenges above, the pipeline successfully localized the ground vechicle in $71$\,sec and with an average position error of $7.9$\,m. This error is comparable to the baseline and stereo-Lidar cases ($6.1$\,m and $8.5$\,m), which demonstrates that the framework is invariant to viewpoints and changes in environment objects.

\begin{figure}[t!]
	\centering%
		\centering
		\includegraphics[width=1\columnwidth]{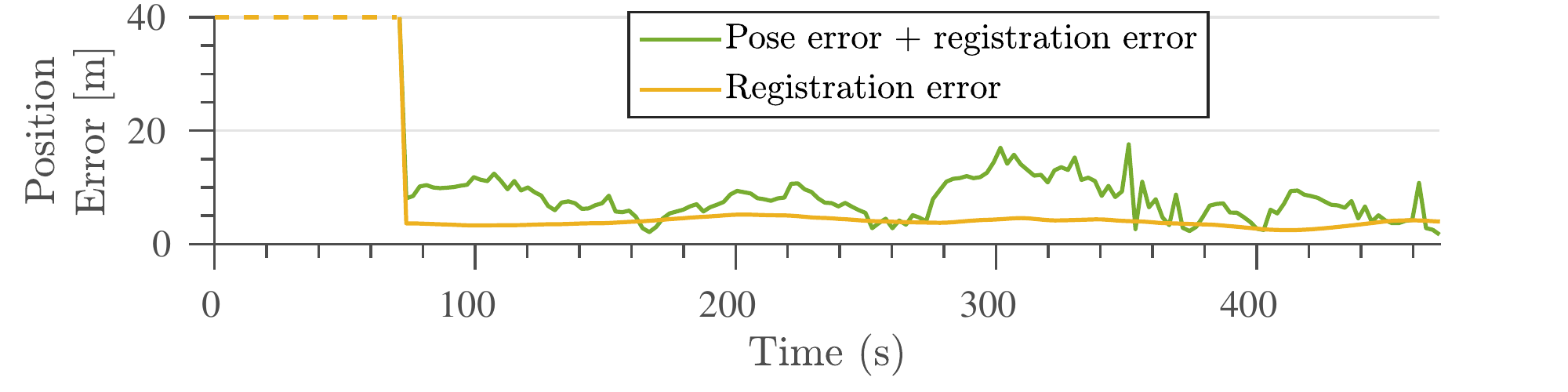}
\vspace*{-4mm}
	\caption{Air-ground localization error in KITTI Sequence 0 by using stereo pose estimates (green) and ground truth pose (yellow). The latter shows the highest attainable localization accuracy, e.g., using a highly-accurate SLAM solution. The occasional lower error from stereo corresponds to instances where the vehicle pose error counteracts the registration error.}
		\vspace*{-0.3em}
	\label{fig:ablation}
\end{figure}

\subsection{Maximum Localization Accuracy}
The accuracy of vehicle pose estimates depends on various factors such as sensors (e.g., camera and Lidar measurements have different noise models that lead to different accuracy), number of loop closures, and robustness of the backend optimizer in the SLAM module.
Pose estimates affect object mapping and egomotion tracking, and therefore directly influence the localization accuracy in our pipeline.
The localization accuracy is also affected by registration. 

Fig. \ref{fig:ablation} analyzes these errors by replacing the ORB-SLAM pose estimates with the the ground truth vehicle pose for localization in the aerial map.
The localization error using the ground truth is only due to registration.
This case has an average position error of $4.1$\,m compared to the $7.9$\,m from stereo pose estimates.
This scenario shows the highest localization accuracy that is achievable by the pipeline, which, for example, can be approached by using a highly-accurate SLAM solution (e.g., top-ranked solutions in the KITTI odometry benchmark \cite{kitti_odom_benchmark}).
The registration error is due to inaccuracies in object centroid estimates in both the ground vehicle map and the aerial reference aerial map (our aerial map has an accuracy of at best $2$\,m).

\subsection{Comparison with Prior Art}

Table \ref{tbl:benchmarks} lists the evaluation results for the real-time localization of a car with an onboard stereo camera (KITTI dataset) in georeferenced aerial images shown in Fig.~\ref{fig:aerial_localization}.
Table \ref{tbl:compare} compares these results with prior art that similarly used the KITTI benchmark for air-ground localization.
The results are carried over here as reported in the corresponding papers since open-source code was not available.
We report the localization error for both cases of stereo odometry and the ground truth odometry, which provides the maximum achievable accuracy of our pipeline.

While the prior state-of-the-art achieves remarkable accuracy, it is unclear if/how these techniques perform in the highly changed environments since the semantic information used (e.g., buildings, road maps, traffic signs) did not change considerably across views in their evaluations. 
In contrast, the challenging scenario created by using cars in an aerial map captured on a different date demonstrates the robustness of our approach in a very different environment leading to comparable localization accuracy in real time.

\begin{table}[t] %
\scriptsize
\centering
\caption{
	Aerial-ground localization comparison on KITTI benchmark for position error and localization time. Dashed line ``$-$'' indicates not reported or not localized successfully.
}
\vspace*{-0.3em}
\setlength{\tabcolsep}{2.3pt}
\begin{tabular}{l c c c c c c c c c c c c }
	\toprule
	\multirow{3}{*}{Approach} & 
	\multirow{3}{*}{\makecell{Ground\\ Modality}} && 
	\multirow{3}{*}{\makecell{Aerial\\ Modality}} &&
	\multicolumn{2}{c}{Seq 0} &&
	\multicolumn{2}{c}{Seq 2} &&
	\multicolumn{2}{c}{Seq 9} \\ 
	\cmidrule{6-7}\cmidrule{9-10} \cmidrule{12-13}
	 &  &&  
	 && \tiny{\makecell{Error\\$[$m$]$}} & \tiny{\makecell{Time\\$[$sec$]$}}  
	 && \tiny{\makecell{Error\\$[$m$]$}} & \tiny{\makecell{Time\\$[$sec$]$}} 
	 && \tiny{\makecell{Error\\$[$m$]$}} & \tiny{\makecell{Time\\$[$sec$]$}} \\
	\toprule
	Fervers \cite{fervers2022continuous} &  L+C  && OP  &&  $-$&$-$  && $1.42$ & $-$  &&  $-$&$-$   \\
	Miller  \cite{miller2021any}  & L+C && OP  &&  $2.0$ & $54.6$ && $9.1$ & $71.5$ && $7.2$ & $75$ \\
	Yan \cite{yan2019global} & L && OSM &&  $20$ & $60$ && $-$&$-$ && $25$ & $17$ \\
	Kim \cite{kim2019fusing} & L && OP &&  $4.6$ & $62$ && $-$&$-$ && $7.7$ & $55.6$ \\
	Brubaker \cite{brubaker2013lost} & S && OSM && $2.1$ &$-$ && $4.1$ & $-$ && $4.2$ & $-$ \\
	Floros \cite{floros2013openstreetslam} & C && OSM && $>10$ & $-$ && $>20$ & $-$ && $-$ & $-$  \\
	\textbf{Ours} & GT && OP &&  $3.9$ & $71$  &&  $9.1$ & $136$   &&  $7.4$ & $86$  \\
	\textbf{Ours} & S  && OP &&  $7.9$ & $71$  &&  $10.6$ & $183$  &&  $10.9$ & $108$ \\
	\bottomrule
\end{tabular}
\label{tbl:compare}
\\ [0.2em]
C: Monocular Camera, ~ S: Stereo Camera, ~ L: Lidar, ~ GT: Ground Truth \\
OP: Orthophoto, ~ OSM: OpenStreetMap
\vspace*{-0.5em}
\end{table}

\section{DISCUSSION} \label{sec:remakrs}

The following discusses some practical limitations and possible extensions of the framework.
1) The localization framework is based on registering geometrically consistent objects, which must be detectable in the vehicle and reference viewpoints.
Thus, while in theory the localization framework is view-invariant, in practice, viewpoint variations can be constrained by the limitations of object classifiers.
2) Environments with repetitive/symmetrical object patterns can create scenarios with a large number of geometrically consistent false matches (e.g., a long line of cars can register to another line on a different street). 
While repetitive patterns (perceptual aliasing) are a fundamental issue in place recognition/localization, steps can be taken to alleviate the problem. For example, object types which have a more random pattern in the environment can be used.
3) Geometric consistency implies that the object maps must have a well-defined scale, which is not the case in monocular SLAM. It is possible to 
extend the formulation using scale-free metrics (such as geometric similarity) at the cost of higher computational complexity.
4) We used a point cloud representation of object centroids, which had sufficient accuracy for car-based localization. However, this representation may not be sufficient for larger objects such as buildings. Hence, other geometric representations such as lines or planar patches should be used instead, and the formulation of geometric consistency can be extended to these representations using proper invariants (cf. prior works \cite{lusk2021clipper, lusk2022global}).

\section{CONCLUSIONS}\label{sec:conclusion}
We presented a view-invariant framework for real-time localization and continuous tracking of a vehicle's pose using compact semantic object maps. Experiments localizing a ground vehicle in reference maps created by stereo cameras and Lidar scans from ground viewpoints, and annotated georeferenced aerial images captured in a different year demonstrated the main contributions of the framework as a \textit{view-invariant} approach robust to \textit{environment object changes}.

\balance %

\bibliographystyle{IEEEtran}
\bibliography{refs}

\end{document}